%% file: main.tex
\documentclass[sigconf, screen]{acmart}
\usepackage{multirow}
\usepackage{enumitem}
\usepackage{svg}
\usepackage{caption}
\usepackage{subcaption}
\usepackage{esvect}
\usepackage{setspace}
\usepackage{float}
\usepackage{colortbl}
\AtBeginDocument{%
  }

\setcopyright{acmlicensed}
\copyrightyear{2024}
\acmYear{2024}
\acmDOI{XXXXXXX.XXXXXXX}

\acmConference[GeoAnomalies'24]{The 1st ACM SIGSPATIAL International Workshop on Geospatial Anomaly Detection}{October 29, 2024}{Atlanta, GA, USA}
\acmISBN{978-1-4503-XXXX-X/18/06}

\acmSubmissionID{216}

\begin{document}

\title[Back to Bayesics: Uncovering Human Mobility Distributions and Anomalies]{Back to Bayesics: Uncovering Human Mobility Distributions and Anomalies with an Integrated Statistical and Neural Framework}


\author{Minxuan Duan}
\affiliation{%
  \institution{Novateur Research Solutions}
  \city{Ashburn}
  \state{Virginia}
  \country{USA}
}
\email{mduan@novateur.ai}

\author{Yinlong Qian}
\affiliation{%
  \institution{Novateur Research Solutions}
  \city{Ashburn}
  \state{Virginia}
  \country{USA}
}
\email{yqian@novateur.ai}

\author{Lingyi Zhao}
\affiliation{%
  \institution{Novateur Research Solutions}
  \city{Ashburn}
  \state{Virginia}
  \country{USA}
}
\email{lzhao@novateur.ai}

\author{Zihao Zhou}
\affiliation{%
  \institution{University of California, San Diego}
  \streetaddress{}
  \city{La Jolla}
  \state{California}
  \country{USA}
}
\email{ziz244@ucsd.edu}

\author{Zeeshan Rasheed}
\affiliation{%
  \institution{Novateur Research Solutions}
  \city{Ashburn}
  \state{Virginia}
  \country{USA}
}
\email{zrasheed@novateur.ai}

\author{Rose Yu}
\affiliation{%
  \institution{University of California, San Diego}
  \streetaddress{}
  \city{La Jolla}
  \state{California}
  \country{USA}
}
\email{roseyu@ucsd.edu}

\author{Khurram Shafique}
\affiliation{%
  \institution{Novateur Research Solutions}
  \city{Ashburn}
  \state{Virginia}
  \country{USA}
}
\email{kshafique@novateur.ai}

\renewcommand{\shortauthors}{Duan et al.}

\begin{abstract}
Existing methods for anomaly detection often fall short due to their inability to handle the complexity, heterogeneity, and high dimensionality inherent in real-world mobility data. In this paper, we propose DeepBayesic, a novel framework that integrates Bayesian principles with deep neural networks to model the underlying multivariate distributions from sparse and complex datasets. Unlike traditional models, DeepBayesic is designed to manage heterogeneous inputs, accommodating both continuous and categorical data to provide a more comprehensive understanding of mobility patterns. The framework features customized neural density estimators and hybrid architectures, allowing for flexibility in modeling diverse feature distributions and enabling the use of specialized neural networks tailored to different data types. Our approach also leverages agent embeddings for personalized anomaly detection, enhancing its ability to distinguish between normal and anomalous behaviors for individual agents. We evaluate our approach on several mobility datasets, demonstrating significant improvements over state-of-the-art anomaly detection methods. Our results indicate that incorporating personalization and advanced sequence modeling techniques can substantially enhance the ability to detect subtle and complex anomalies in spatiotemporal event sequences.
\end{abstract}




\maketitle

\begin{spacing}{0.975}

\input{introduction.tex}
\input{preliminary}

\input{method.tex}

\input{experiment.tex}

\input{conclusion.tex}
\end{spacing}


\begin{acks}
Research supported by the Intelligence Advanced Research
Projects Activity (IARPA) via the Department
of Interior/Interior Business Center (DOI/IBC) contract
number 140D0423C0033. The U.S. Government is authorized
to reproduce and distribute reprints for Governmental
purposes, notwithstanding any copyright annotation
thereon. Disclaimer: The views and conclusions contained
herein are those of the authors and should not
be interpreted as necessarily representing the official
policies or endorsements, either expressed or implied, of
IARPA or the U.S. Government.
\end{acks}

\newpage
\bibliographystyle{ACM-Reference-Format}
\bibliography{base}

\appendix 
\newpage


\end{document}

%% file: introduction.tex
\section{Introduction}
How can we gain insights from tracking the movement of human populations? The exploration of human mobility, encompassing its dynamics, causes, motivations, and limitations, has been a focus of study since at least the 19th century, with foundational contributions such as Ravenstein's laws of migration \cite{ravenstein1885laws}. In contemporary research, this field has advanced significantly through the use of various positioning technologies, such as GPS, cellular networks, and WiFi \cite{zheng2013effective, gonzalez2008understanding, hasan2013spatiotemporal}. These technologies enable the detailed analysis of movement patterns at both individual and societal levels, providing critical information for professionals in urban planning, transportation, and public health monitoring \cite{barbosa2018human, luca2021survey}.

Anomalies in mobility data --- unusual or unexpected patterns of movement --- can signify a range of events, from transportation disruptions to shifts in population behavior caused by emergencies, pandemics, or large public gatherings. Identifying these anomalies enables stakeholders to respond effectively, ensuring public health, optimizing infrastructure, and maintaining urban resilience.

Yet, the analysis of mobility data is fraught with challenges that complicate the detection of anomalies. Mobility data is often sparse, with gaps in coverage and missing data points, making it difficult to establish a clear baseline of normal behavior. Additionally, human movement patterns are highly complex and influenced by a myriad of factors, resulting in multivariate dependencies that traditional models struggle to capture. These complexities, coupled with the dynamic nature of human behavior, make it difficult to distinguish between true anomalies and normal variability in the data. Consequently, traditional statistical methods and machine learning models frequently fail to detect subtle yet significant anomalies.

To effectively detect anomalies in mobility data, it is essential to first understand the underlying distribution of normal behavior. Recovering these underlying distributions is crucial because it provides a baseline against which deviations can be measured. However, this task is inherently challenging due to the intricate nature of human mobility, which is influenced by a wide array of variables such as time, agent identity (e.g., individual or vehicle IDs), points of interest, and social and environmental factors.

For instance, consider the problem of predicting traffic patterns in a large city. The movement of vehicles is not only dependent on time of day and location but also on events such as road closures, public holidays, or weather conditions. Additionally, the identity of the agent (e.g., individual or vehicle performing activities) is crucial in understanding normal versus anomalous behavior. For example, a late-night drive through a typically congested downtown area might be considered anomalous for a typical commuter but perfectly normal for a taxi driver whose work involves frequent trips at all hours. Similarly, what is considered a normal pattern of movement for a door-to-door salesperson would be highly anomalous for someone who typically works from home. These examples underscore the necessity of capturing the full complexity of mobility data, which includes understanding how the variables such as time, location, and agent ID interact to form the overall distribution of normal behavior~\cite{shsu2024trajgpt}.

Existing methods for recovering these underlying distributions often fall short because they struggle to manage the complexity, heterogeneity, and high dimensionality of mobility data (See Section 2 for a detailed discussion). To address these challenges, many models rely on simplifying assumptions, such as treating variables as independent. However, these assumptions rarely hold true in real-world mobility data, where variables such as time and location are deeply interdependent. As a results these models frequently produce incomplete or inaccurate representation of the underlying distributions, overlooking critical nuances in the data. 

In response to these challenges, we propose DeepBayesic, a novel framework that integrates Bayesian principles with advanced neural density estimation techniques to recover the underlying multivariate distributions from sparse and complex mobility datasets. Our approach is designed to handle the high dimensionality, heterogeneity, and interdependencies inherent in mobility data, providing a more accurate and comprehensive understanding of normal behavior. It combines the complementary strengths of Bayesian theory and neural networks: Bayesian theory offers a probabilistic framework for integrating prior knowledge and managing uncertainty, while neural networks are well-suited for capturing complex, high-dimensional relationships within the data. 

DeepBayesic employs a cascade of neural density estimators to model the complex interactions between heterogenous variables such as time, location, and agent ID. By using a Bayesian framework, the model captures the full multimodal distribution of possible outcomes, rather than relying on single-point predictions. This allows DeepBayesic to identify anomalies that are not only deviations from the expected outcome but also deviations from the entire range of normal behavior.

To personalize the model for individual agents (e.g., specific vehicles, individuals, or groups of individuals with similar socio-economic demographics), DeepBayesic incorporates agent embeddings. These embeddings capture the unique characteristics and behaviors of different agents, allowing the model to better distinguish between normal and anomalous behavior on an individual or group basis. This personalized approach enhances the model's sensitivity to subtle anomalies that might otherwise go unnoticed.

We demonstrate the effectiveness of DeepBayesic through extensive experiments on multiple mobility datasets, showing that it outperforms existing anomaly detection methods. 

Our contributions can be summarized as follows:
\begin{itemize}
    \item We introduce DeepBayesic, a novel framework that combines Bayesian theory with deep neural networks to accurately recover underlying multivariate distributions from sparse and complex mobility data. 
    \item We implement a cascade of neural density estimators that can handle the high dimensionality and complex interdependencies in mobility data, capturing both simple and complex anomalies. 
    \item We develop and incorporate deep agent embeddings to personalize the model, improving its ability to detect anomalies specific to individual agents. 
    \item We demonstrate the effectiveness of DeepBayesic through extensive experiments on multiple mobility datasets, showcasing its superiority over existing baselines in anomaly detection. 
\end{itemize}

The remainder of this paper is organized as follows. In Section \ref{sec:related work}, we discuss the innovations presented in this work within the context of the state of the art in anomaly detection and neural density estimation, highlighting the key advancements that distinguish our approach. Section \ref{sec: problem statement} formulates the problem, providing a formal definition and setting the stage for the methodological developments that follow. Section \ref{sec: method} details the methodology, describing the structure and components of the proposed DeepBayesic framework, including the integration of Bayesian principles, neural density estimation, and agent embeddings. In Section \ref{sec: experiments}, we present extensive experiments conducted on multiple mobility datasets to evaluate the effectiveness of our approach, comparing its performance with existing baselines. Finally, Section \ref{sec:conclusion} concludes the paper, summarizing the key findings and discussing potential directions for future research.

\section{Relationship with State-of-the-Art}\label{sec:related work}
In this section, we position our proposed framework within the broader landscape of existing research in the areas of anomaly detection (Section \ref{sec:related work-anomaly detection}) and neural density estimation (Section \ref{sec:related work-neural density estimation}).  
\subsection{Anomaly Detection}\label{sec:related work-anomaly detection}
Anomaly detection in spatiotemporal data has garnered significant attention, particularly for identifying irregular patterns in user behavior. Existing approaches to anomaly detection in mobility data can be broadly categorized into clustering-based, distance-based, reconstruction-based, prediction-based, and density estimation-based methods. \textit{Clustering-based methods} group data into clusters and flag anomalies as outliers ~\cite{li2021clustering, pu2020hybrid,kiss2014data}. These methods often assume that normal data forms distinct clusters, an assumption that may not hold true in complex mobility data. \textit{Distance-based methods} detect anomalies by measuring deviations from the norm ~\cite{arias2023aida,sarafijanovic2019fast,huo2019anomalydetect,liu2024neuralcollaborativefilteringdetect}. While intuitive and straightforward to implement, these approaches heavily rely on the choice of distance metric --- a challenge in complex datasets where different features may have varying scales or significance. Moreover, they suffer from the curse of dimensionality, where the effectiveness of distance metrics diminishes as the number of dimensions increases, making it difficult to distinguish between normal and anomalous points. \textit{Reconstruction-based methods} use models like deep autoencoders to reconstruct data and identify anomalies based on reconstruction errors. Although these methods have shown promise ~\cite{zhou2017anomaly, liu2020online}, they are prone to overfitting, where the model may learn to perfectly reconstruct the training data but fail to generalize to new, unseen data. Additionally, they are sensitive to outliers in the training data, which can lead the model to incorrectly learn to reconstruct these anomalies as normal patterns, thereby diminishing its effectiveness in identifying true anomalies. 

\textit{Prediction-based methods} have been a major focus of anomaly detection research ~\cite{lam2016concise, wu2017fast, keane2017detecting, irvine2018normalcy, song2018anomalous, liu2020online}. These methods forecast future data points and identify anomalies as deviations from these predictions. However, existing predictive models~\cite{de2015artificial, zhang2018multi, rossi2019modelling, zhang2019prediction, li2020fast, ebel2020destination, liao2021taxi, hoch2015ensemble, gupta2018taxi, fu2019deepist, lan2019travel, das2019map, abbar2020stad, mobtcast, starhit} have severe limitations when forecasting complex mobility patterns. Typically, these models predict only the location of visits~\cite{mobtcast, starhit}, which limits their utility in detecting temporal anomalies. While some recent methods have attempted to model both location and time \cite{deepjmt, getnext}, they often make limiting assumptions, such as the independence between location and time, which can affect their predictive accuracy~\cite{shsu2024trajgpt}. 

One of the major drawbacks of prediction-based models is their reliance on single-point predictions, which can overlook the multimodal nature of real-world data distributions. These models typically generate a single expected outcome, assuming that future behavior can be represented by a specific predicted value. However, in reality, the future may involve multiple plausible outcomes, each with its own probability. For example, in urban mobility, travel time might vary widely due to factors like traffic conditions or weather, leading to a multimodal distribution of possible travel times. Single-point predictions can average out these possibilities, resulting in inaccurate forecasts and the potential to miss critical anomalies that arise from these alternative scenarios.

This limitation of prediction-based models underscores the need for approaches that can capture the full distribution of possible outcomes rather than just a single expected value. \textit{Density estimation-based methods} address this by modeling the entire probability distribution of the data and identifying anomalies as data points that fall in regions of low probability. These methods are better suited for detecting a broader range of anomalies, including those arising from complex, multimodal distributions. However, despite their advantages, density-based approaches also face challenges. Traditional methods, such as kernel density estimation (KDE), can struggle with high-dimensional data due to the curse of dimensionality, where the volume of the data space increases exponentially with the number of dimensions, making it difficult to estimate densities accurately. Moreover, density estimation methods often require large amounts of data to model the joint distribution effectively, which can be a limitation in scenarios with sparse data or missing values. Additionally, these methods may require careful tuning to balance the trade-off between bias and variance, where overly smooth density estimates may miss subtle anomalies, while overly complex models may overfit the data. 




\subsection{Neural Density Estimation}\label{sec:related work-neural density estimation}
Neural networks are widely recognized as universal function approximators, making them highly effective tools for modeling complex systems. For mobility data analysis, neural density estimators harness this expressive power to model sequences of multivariate distributions and stochastic processes, effectively capturing the intricate patterns of human movement in high-dimensional spaces. These estimators are especially suited for representing the temporal and spatial dependencies inherent in mobility data, where each observation can be treated as a sample from a time-varying distribution. Neural density estimators for mobility analysis can be categorized based on their approach for handling temporal dynamics and their specific techniques for modeling probability densities.

Continuous-time models offer the advantage of capturing the fluid nature of mobility patterns. \textit{Neural Ordinary Differential Equations} (Neural ODEs)~\cite{chen2018neural} model the continuous evolution of the system state over time, allowing for the hidden state formulation between two events. \textit{Neural Stochastic Differential Equations} (Neural SDEs)~\cite{jia2019neural} build on this approach by incorporating stochastic elements, which more effectively capture the inherent randomness in mobility patterns. Another continuous-time approach, is \textit{Neural Point Processes}~\cite{mei2017neural}, which excels at modeling the occurrence rates of sporadic mobility events in continuous time. These models are capable of estimating temporal densities and may offer tractable likelihood computation~\cite{zhou2022neural}, making them particularly suitable for density estimation tasks in mobility data analysis. In contrast, discrete-time models, such as Autoregressive models~\cite{papamakarios2017masked}, Recurrent Neural Networks (RNNs) and Transformer-based architectures~\cite{transformer}, operate on fixed time steps. These discrete-time approaches can be effectively combined with various density estimation techniques to provide explicit likelihood computations.

The spatial dimension of mobility data poses unique challenges for density estimation. \textit{Normalizing flows}\cite{rezende2015variational, kobyzev2020normalizing} have shown promise in modeling complex spatial distributions, offering both flexible modeling capabilities and exact likelihood computation. However, they can be computationally intensive, especially in high-dimensional spaces. To address the diverse features often present in mobility data, various approaches have been proposed. \textit{Mixture Density Networks}~\cite{bishop1994mixture}, such as a mixture of Gaussian or Laplace densities~\cite{uria2013rnade, tenzer2024generating}, can be used as output layers to enable modeling of complex, multimodal distributions at each time step. \textit{Variational Autoencoders} (VAEs)~\cite{kingma2013auto} learn latent representations of the data and provide a lower bound on the likelihood. Another approach, \textit{Automatic Integration}~\cite{zhou2024automatic}, learns the density directly as a neural network, offering faster computation than normalizing flows in low dimensional settings, but it does not scale effectively to higher dimensional outputs. Recent advancements in \textit{Diffusion Models}~\cite{ho2020denoising} also show promise, as they can be adapted to provide likelihood estimates through variants like Variational Diffusion~\cite{kingma2021variational}. The choice of model often requires balancing expressiveness against the computational tractability of likelihood estimation. 

The proposed DeepBayesic framework is closely related to autoregressive models and mixture density networks discussed above, but it offers several key advantages over existing approaches. First, it excels in handling heterogeneous inputs by seamlessly integrating both continuous and categorical data. While most autoregressive models primarily focus on modeling distributions of either discrete tokens (e.g., LLMs or grid-based locations~\citep{liu2020online}) or continuous output~\citep{zhou2022neural}, they often often struggle to handle both types of data simultaneously. This limitation poses significant challenges in real-world applications, such as mobility data analysis, where categorical inputs --- such as types of points of interest (POI) --- provide crucial context and are essential for accurate modeling. Another critical advantage of DeepBayesic lies in its tailored modeling of diverse feature distributions. In many datasets, different features follow significantly different distributions. For example, arrival times might follow a multimodal normal mixture, while duration could adhere to a log-normal distribution with a strong left skewness. Many autoregressive models treat all features uniformly, which can lead to oversights and inaccuracies by ignoring these important distributional nuances. DeepBayesic addresses this challenge by allowing for the customization of neural density estimators tailored to the specific characteristics of each conditional distribution. Unlike existing autoregressive models, which often rely on a shared architecture across all features, DeepBayesic enables the integration of different types of neural networks, such as recurrent networks for modeling point-of-interest sequences and transformer networks for handling stay durations. This modular approach enables optimization of the architecture for the specific requirements of each data type. By customizing each estimator to capture the unique complexities of the data, DeepBayesic provides a more robust and adaptable framework for mobility data analysis.

%% file: preliminary.tex
\section{Problem Statement}\label{sec: problem statement}
Our study focuses on finding anomalies in large-scale urban mobility sequences. A mobility sequence represents a series of recorded movements or activities of an agent, such as an individual or a vehicle, over time.Let $P$ denote the total number of agents, where each agent $z$, indexed by $z\in \{1, \ldots, P\}$, is observed through a mobility sequence denoted by $\boldsymbol{X} = [X_{1},\ ..., X_{n}]$. Each observation $X_{i}$ corresponds to a specific event or activity characterized by a set of attributes such as speed, location, time, and the type of visited place. Formally, each observation $X_{i}$ is represented as:
$X_{i}=(a_{i,1}, a_{i,2}, a_{i,3}, \ldots, a_{i,m})$
where $a_{i,j}$ denotes the $jth$ attribute of $X_{i}$. Anomaly detection is achieved by estimating the probability distribution of these observations and identifying instances that significantly deviate from the estimated distribution. 

To estimate the probability distribution of the observation $\boldsymbol{X}$ for a given agent $z$, denoted as $P(\boldsymbol{X}|z)$, we aim to estimate the joint distribution of the attributes that constitute $\boldsymbol{X}$. By applying Bayesian inference and the chain rule of probability, the probability distribution $P(\boldsymbol{X}|z)$ can be decomposed into a product of conditional probabilities:

\begin{equation}
\begin{aligned}
P(\boldsymbol{X} \mid z) &= P(a_{:,1}, a_{:,2}, \ldots, a_{:,j} \mid z) \\
&= P(a_{:,1} \mid z) \times P(a_{:,2} \mid a_{:,1}, z) \times \cdots \\
&\quad \times P(a_{:,j} \mid a_{:,1}, a_{:,2}, \ldots, a_{:,j-1}, z)
\end{aligned}
\label{eq:chain rule general}
\end{equation}

Each term in this sequence of conditional probabilities can be estimated using a suitable density estimator, enabling us to capture the complex dependencies between attributes effectively. 

In this study, we concentrate on three key attributes for each observation that are particularly informative for identifying deviations from typical behavior patterns: arrival time $a_{i,1}:=t_{i}$, stay duration $a_{i,2}:=d_{i}$, and POI type $a_{i,3}:=c_{i}$. Thus, each observation $X_i$ of agent $z$ is defined as:
\begin{equation}
    X_i = (c_i, t_i, d_i)
\end{equation}

For each agent $z$ in our dataset, there is an associated mobility sequence, $\boldsymbol{X_{train}}$, which predominantly consists of normal activities. These sequences collectively constitute the training dataset used to learn the underlying probability distributions in an unsupervised fashion. Our goal in the anomaly detection task is to assign an anomaly score $s$ to each observation in a separate set of mobility sequences, $\boldsymbol{X_{test}}$. According to Eq. \ref{eq:chain rule general}, the probability distribution $P(\boldsymbol{X}\ |\ z)$ can be rewritten as:

\begin{align}
\begin{split}
P(\boldsymbol{X} \mid z) 
&= P(d,\ c,\ t  \mid z) \\
&=  P(t \mid z)\ 
    P(c\ \mid t,\ z)\ 
    P(d  \mid c,\ t,\ z)
\end{split}
\label{eq:bayes}
\end{align}

Given an estimate of the probability distribution $P(\boldsymbol{X^{train}}\ |\ z)$, the anomaly score for each observation in $\boldsymbol{X_{test}}$ can be determined by calculating $1 - P(\boldsymbol{X^{test}}|z)$.

%% file: method.tex
\section{Method}\label{sec: method}
Given an agent $z$, estimating the probability distribution of the associated mobility sequence $\boldsymbol{X_{train}}$ is challenging due to the typically sparse data available for individual agents. To address this challenge, we employ an agent embedding model to extract latent features from the observed activities of all agents, clustering those with similar behaviors closer together in the latent space. This approach enables us to estimate $P(\boldsymbol{X} \mid h)$, where $h$ is the learned agent embedding, rather than directly estimating $P(\boldsymbol{X} \mid z)$ from sparse data.
Following Eq. \ref{eq:bayes}, given an agent embedding $h$, we estimate the probability distribution of each observed sequence, $\boldsymbol{X}$, by sequentially estimating the conditional probability distributions of its attributes, the arrival time $(P(t \ |\ h))$, the type of point of interest $(P(c\ |\ t, h))$, and the stay duration $(P(d\ |\ c, t, h))$. 

The subsequent sections will detail the process of obtaining agent representations and the implementation of 
the three conditional probability distributions.


\subsection{Agent Embedding Auto-Encoder}\label{sec: agent embedding}
\begin{figure}
    \centering
    \includegraphics[width=0.9\linewidth]{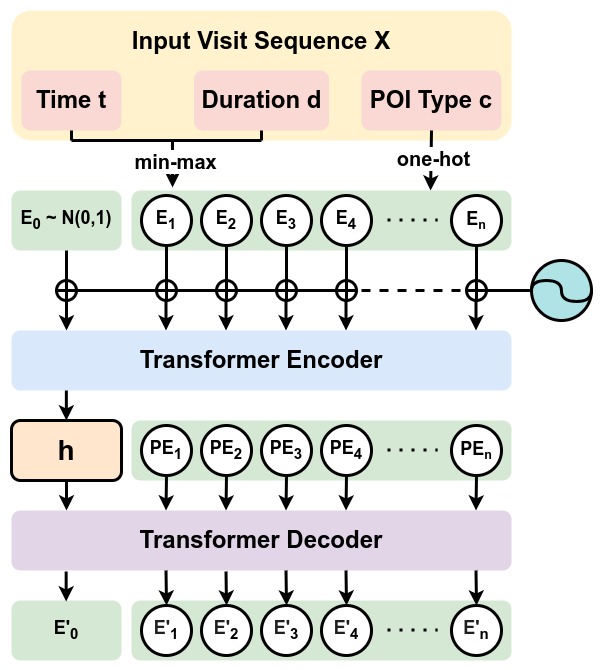}
    \caption{Agent embedding auto-encoder. A Transformer Encoder is trained to project an input feature sequence $\boldsymbol{X}$, encoded by a Sequence Encoder and prefixed by learnable token $E_{0}$, into its latent representation $h$. Simultaneously, a Transformer Decoder is trained to reconstruct the encoded sequence from the latent representation $h$ and the standard positional encoding $\boldsymbol{PE}$.}
    \label{fig:agent_embed_block}
\end{figure}

 Fig. \ref{fig:agent_embed_block} illustrates the architecture of the agent embedding auto-encoder. We employ a transformer-based auto-encoder, similar to MotionClip ~\cite{tevet2022motionclip}, to map each staypoint sequence to an agent embedding vector. This model is trained to project the staypoint sequence $\boldsymbol{X}$ of an agent $z$ into a latent vector $h$ (the agent embedding vector) while simultaneously reconstructing the original sequence. The details of each component of this agent embedding auto-encoder are provided in the following subsections. 
\subsubsection{Sequence Encoder}\label{sec:sequence encoding}
We begin by encoding each staypoint $X_i=(c_i, t_i, d_i)$ into an encoded space $E_i$ to normalize the data representation. Specifically, we use one-hot encoding for the POI type and use min-max normalization for arrival time and stay duration. A learnable token sequence $E_{0}$, sampled from standard normal distribution, is inserted at the beginning of the encoded sequence, following the approach used in ~\cite{tevet2022motionclip}. The input to the transformer encoder is denoted as $E = [E_{0}, E_{1},\ ..., E_{n}]$. Further details on the encoding process can be found in Section \ref{sec:preprocessing}. 
\subsubsection{Transformer Encoder}
The transformer encoder maps the encoded staypoint sequence $E$ to a latent representation $h$. First, the encoded sequence is projected into the encoder's dimension by linear projection. Next, a standard positional embedding is applied to the projected sequence. The latent representation $h$ is then obtained as the first output of the encoder, with the remaining sequence outputs discarded. 
\subsubsection{Transformer Decoder}
The latent representation $h$ is fed to the transformer decoder as key and value, while the positional encoding is inputted to the transformer decoder as query sequence. Subsequently, the transformer decoder predicts an encoded staypoint sequence $E'$.
\subsubsection{Loss Function of the Agent Embedding Auto-Encoder}
 The agent embedding auto-encoder is trained by minimizing the reconstruction loss, defined as the mean squared error between the input sequence $E$ and the reconstructed output sequence $E'$. Specifically, the loss function is given by:

\begin{equation}
\mathcal{L}_{\text{ae}} = \frac{1}{n*|E|} \sum_{i=1}^{n} \lVert E_{i}-\hat{E_{i}} \rVert^2
\label{eq:agent embedding loss}
\end{equation}

\subsection{Joint Distribution Estimation}
\begin{figure*}
    \centering
     \includegraphics[width=0.9\linewidth]{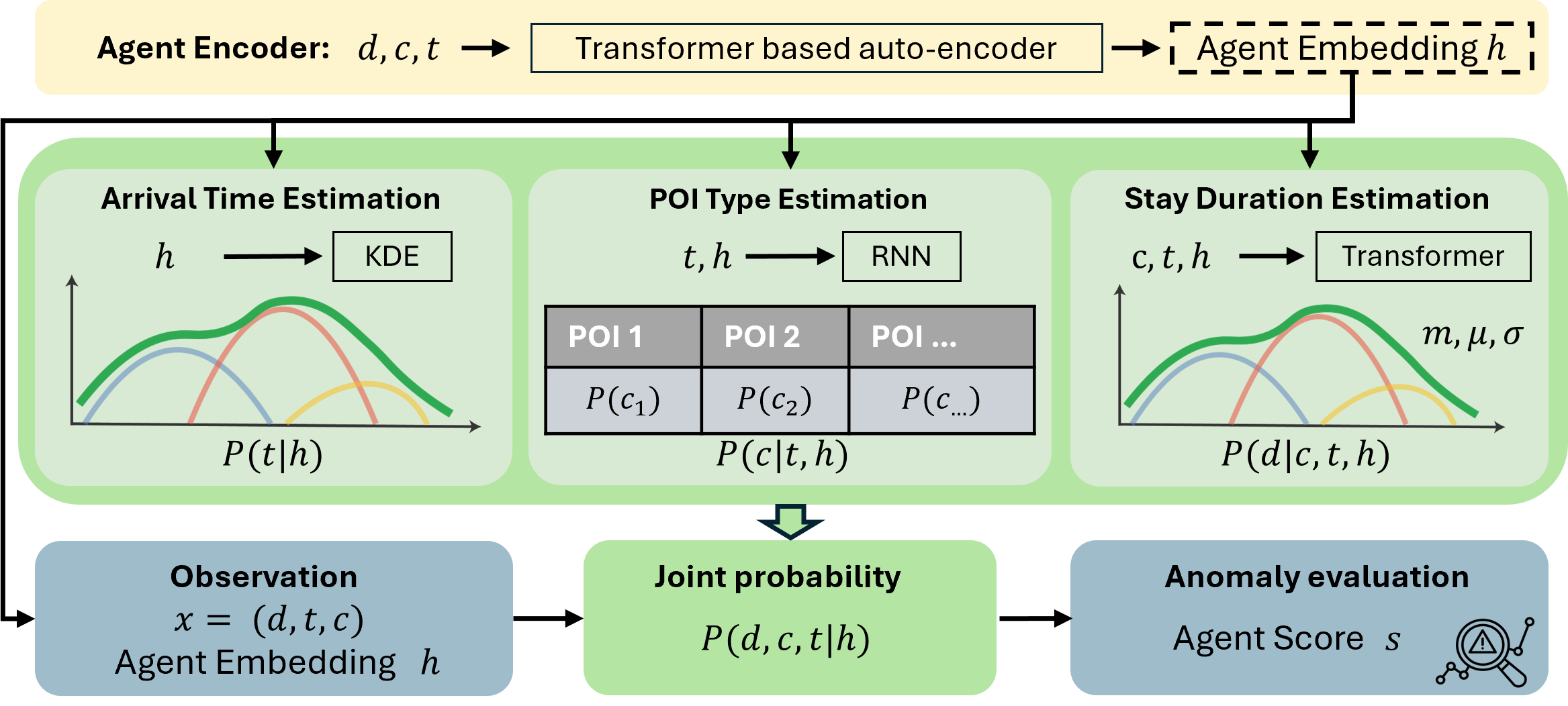}
    \caption{DeepBayesic pipeline. The agent embedding $h$ is fed into three modules: Arrival Time Estimation, POI Type Estimation, and Stay Duration Estimation. Each module uses $h$ along with other relevant inputs to estimate the corresponding conditional probability distributions, which are then integrated into a joint probability model. Finally, the agent embedding $h$ and an observation $x$ are input into the joint probability model to calculate anomaly score $s$.}
    \label{fig:autoSTM.png}
\end{figure*}

In this section, we provide a detailed explaination of the estimation process for the joint probability distribution $P(X\ |\ h)$, as outlined in Section \ref{sec: problem statement}. The pipeline for this process is illustrated in Fig. \ref{fig:autoSTM.png}.

\subsubsection{Arrival Time Probability Estimation}

To estimate the probability distribution of arrival times, $\hat{P}(t|h)$, for a given agent, we employ a kernel-based Gaussian mixture model (GMM), denoted as $f_{t}(h)$. This model is trained on the arrival time sequence observed during the training period. The estimated probability is defined as: 
\begin{equation}
    \hat{P}(t\ |\ h)\ :=\ f_{t}(h)
\end{equation}

Since arrival times are inherently non-negative, a clipping function is applied to ensure that all estimated arrival times remain within valid bounds.

\subsubsection{POI Type Probability Estimation}
Given the agent embedding $h$ and arrival time $t$, we use a Recurrent Neural Network (RNN), denoted by $f_{POI}$, to estimate the probability distribution over the POI type $c$. We chose an RNN architecture because it is particularly effective at handling sequential data and capturing temporal dependencies. After obtaining the hidden state vector $\mathbf{g}=(g1, \ldots, g_n)$ from the RNN, we apply a linear projection followed by a softmax function to $\mathbf{g}$ to predict the probability distribution over the discrete POI types:
\begin{equation}
    \hat{P}(c\ |\ t,\ h\ ) = \text{softmax}(\mathbf{W} \mathbf{g} + \mathbf{b})
\end{equation}
where $\mathbf{W}$ is the weight matrix and $\mathbf{b}$ is the bias vector of the linear projection layer. The softmax function is defined as:
$$
\text{softmax}(y_i) = \frac{\exp(y_i)}{\sum_{j=1}^{K} \exp(y_j)}
$$
where $y_i$ is the $i$-th element of the output vector, and $K$ is the number of POI types.




\subsubsection{Stay Duration Probability Estimation}
Given the agent embedding $h$, arrival time $t$, and POI type $c$, we use a transformer-based neural density function, $f_d$, to model the probability distribution of stay duration $d$. As discussed in Section \ref{sec:related work-neural density estimation}, a neural density function $f$ is a neural network that implicitly parameterizes the variables to be estimated. This approach is especially well-suited for handling high-dimensional datasets with sparse samples, a common characteristic of activity sequence data. When combined with Bayesian inference, it offers a robust framework for incorporating and updating prior knowledge as more data attributes are observed. 

To model the distribution of stay duration, $\hat{P}(d\ |\ c,\ t,\ h)$, we represent it using a Gaussian Mixture Model (GMM), which can be expressed as:

\begin{equation}
    \hat{P}(d\ |\ c,\ t,\ h)\ := \sum_k m_k p_{\mathcal{N}}(d; \mu_k, \sigma_k)
\end{equation}
where:
\begin{equation}
    p_{\mathcal{N}}(d; \mu_k, \sigma_k) = \frac{1}{\sigma_k \sqrt{2 \pi}} \exp \left( -\frac{ (d-\mu_k)^2 }{2 \sigma_k^2} \right)
\end{equation}

The mixture weights $\boldsymbol{m}$ are estimated by the neural density function: 
\begin{equation}
    f_{d}: c,\ t,\ h \rightarrow \mathbf{m}
\end{equation}

A softmax function is applied to ensure that the mixture weights $\boldsymbol{m}$ sum to 1. For a given observation of stay duration $d$, the likelihood under the model $p_{GM}$ is maximized by minimizing its negative log-likelihood, forming the basis for the training loss function for $f_{d}$:
\begin{align}
\begin{split}
    \mathcal{L}_{f_{d}}(d) 
                   &= -\log \left[\sum_k m_k p_{\mathcal{N}}(d; \mu_k, \sigma_k)\right]
\end{split}
\label{eq:loss of f}
\end{align}

The transformer-based neural density function facilitates attention mechanisms across $h$, $t$, and $c$, allowing for an adaptive representation that captures complex dependencies between these variables.

\subsubsection{Neural Density Function Loss}
Using the chain rule of probability (Eq. \ref{eq:bayes}), the loss function for training the joint distribution estimation function is defined as:
\begin{equation}
\mathcal{L}_{\text{f}} = -\sum_{c, d}
    \left[\log\hat{P}(c \mid t, h)
    + \log\hat{P}(d \mid c, t, h)\right]
\label{eq:neural-density-function loss}
\end{equation}

\subsubsection{Total loss}
The total loss function is obtained by combining the agent embedding autoencoder loss (Eq. \ref{eq:agent embedding loss}) with the neural density function loss (Eq. \ref{eq:neural-density-function loss}). The total loss is:

\begin{equation}
    \begin{aligned}
        \mathcal{L}_{\text{total}} &= \mathcal{L}_{\text{ae}} + \mathcal{L}_{\text{f}} \\
        &= \frac{1}{n*|E|} \sum_{i=1}^{n} \lVert E_{i}-\hat{E_{i}} \rVert^2 - \sum_{c, d}
        \left[\log\hat{P}(c \mid t, z)
        + \log\hat{P}(d \mid c, t, z)\right]
    \end{aligned}
\end{equation}

\subsection{Anomaly Score Assignment}
Given the agent embedding $h$, the inferred staypoint sequence $X^{infer}$, and probability estimates $\hat{P}(t\ |\ h)$, $\hat{P}(c\ |\ t,\ h)$, and $\hat{P}(d\ |\ c,\ t,\ h)$, we first compute the joint probability $\hat{P}(d^{infer},\ c^{infer},\ t^{infer}\ |\ h)$ as follows:

\begin{equation}
    \begin{aligned}
    \hat{P}(d,\ c,\ t\ |\ h) =\  &\text{CLIP}_{arrival\_time}\left(\hat{P}(t\ |\ h)\right)\times \\
    &\ \text{CLIP}_{POI\_type}\left(\hat{P}(c\ |\ t,\ h)\right)\times \\
    &\ \ \text{CLIP}_{duration}\left(\hat{P}(d\ |\ c,\ t,\ h)\right)
    \end{aligned}
\end{equation}
where $CLIP$ denotes a clipping function applied to ensure valid values. The anomaly score $s$ is then computed as:
\begin{equation}
    s=1-\hat{P}(d,\ c,\ t\ |\ h).
\end{equation}

%% file: experiment.tex
\section{Experiments and Results}\label{sec: experiments}
In this section, we present the experiments conducted to evaluate the performance of the proposed DeepBayesic framework. Our experiments (Section \ref{sec: experimental_setup}) are designed to assess the model’s effectiveness in anomaly detection (Section \ref{sec: anomaly_results}), its ability to personalize predictions for individual agents (Section \ref{sec:personalization}), and the impact of various model components on overall performance (Section \ref{sec:ablation}). 

\subsection{Experimental Setup}\label{sec: experimental_setup}
\subsubsection{Datasets}
To evaluate the performance of our proposed approach, we conducted experiments using two synthetic mobility datasets, NUMOSIM-LA \cite{stanford2024numosimsyntheticmobilitydataset} and Urban Anomalies \cite{amiri2024anomaly,amiri2024patternslifehumanmobility} that simulate human movement patterns in urban environments.

\textbf{NUMOSIM-LA}\footnote{The NUMOSIM-LA dataset can be accessed at: \url{https://osf.io/sjyfr/}.}: NUMOSIM is a synthetic dataset designed to simulate human mobility patterns within urban environments\cite{stanford2024numosimsyntheticmobilitydataset}. The initial release, NUMOSIM-LA, focuses on the Los Angeles area and is divided into four weeks for training and four weeks for testing. It contains data for 200,000 agents, of which 381 exhibit various types of anomalous behaviors during the test period. These behaviors range from isolated events that disrupt an agent's typical sequence of activities to recurring patterns that deviate from the norm regularly. At the staypoint level, the dataset comprise a total of 16,667,273 staypoints, with 3,468 labeled as anomalous. The anomaly prevalence rates are 0.19\% at the agent level and 0.0208\% at the staypoint level. This distribution reflects the significant imbalance typically encountered in real-world anomaly detection problems. Additionally, NUMOSIM-LA provides several benchmarks from state-of-the-art methods, making it a valuable resource for evaluating the effectiveness of the DeepBayesic model against current standards.

\textbf{Urban Anomalies}\footnote{The Urban Anomalies dataset can be accessed at: \url{https://osf.io/dg6t3/}.}: The Urban Anomalies dataset \cite{amiri2024anomaly, amiri2024patternslifehumanmobility} includes synthetic simulations of urban environments in Atlanta and Berlin, each containing mobility sequences for 1000 agents, some of whom are affected by a simulated disease that causes them to eat more frequently, reduce social interactions, and stop going to work. The anomaly prevalence rates are 12\% at the agent level and 12.3\% at the staypoint level. Each mobility sequence includes one month of normal activities followed by a period of variable length with injected anomalies. The anomalies fall into four categories: i) \textit{Hunger}: Agents visit restaurants more frequently, ii) \textit{Social}: Agents visit random locations instead of meeting friends, iii) \textit{Work}: Agents stop going to work, and iv) \textit{Combined}: A mix of all the above anomalies. The dataset provides comprehensive information, including agent trajectories, staypoints, and social links, offering a rich set of features for evaluating anomaly detection methods.


\subsubsection{Preprocessing}
For each dataset, we computed the following features for each staypoint:
\begin{itemize}
    \item Time: Epoch time (day/hour)
    \item Location: Latitude and longitude
    \item Duration: Length of stay
    \item POI type: Type of point of interest
    \label{sec:preprocessing}
\end{itemize}

Epoch time was normalized to center around Monday, combining both day and hour values. Urban Anomalies dataset directly provides POI type as an attribute. For NUMOSIM-LA, we determined POI type of each stay point by mapping it to the nearest available POI within a 15-meter radius, or labeling it as 'unknown' when no POI was identified within this distance. Each baseline model was adapted to support the NUMOSIM and Urban Anomalies datasets (See \cite{stanford2024numosimsyntheticmobilitydataset} for details of these modified pipelines).

\subsubsection{Baseline}

We evaluate the performance of our anomaly detection pipeline against the following baseline methods:

\textbf{RioBusData} \cite{Bessa_Silva_Nogueira_Bertini_Freire_2016}: A convolutional neural network designed to detect outlier trajectories in the bus routes of Rio de Janeiro. The input sequence consists of agent IDs and all the features listed in Section \ref{sec:preprocessing}. Location feature was represented by raw geographic latitudes and longitudes.
The model's output was modified to predict anomalies in agent behaviors instead of bus routes.

\textbf{Spatial-Temporal Outlier Detector (STOD)} \cite{Cruz_Barbosa_2020}: A GRU-based neural network that detects anomalies in bus trajectories using GPS points from regular bus routes. This model utilizes all the features in Section \ref{sec:preprocessing}. An embedding layer was used to encode latitudes and longitudes into tokens. The H3 resolution used in Geo embedding creation was set to 12.

\textbf{Gaussian Mixture Variational Sequence AutoEncoder (GM-VSAE)} \cite{liu2020online}: A VAE-based model designed to detect trajectory anomalies through trajectory generation. The latitudes and longitudes were converted into grid indices, and were used as the only input to the model. This model is limited to detecting anomalies at the agent level, without providing insights into anomalies at the stay-point level.


\subsubsection{Metrics}\label{sec:metrics}
To evaluate the performance of our approach, we use the following metrics:
\begin{itemize}
    \item Area Under the Precision-Recall Curve (AUPR)
    \item Area Under the Receiver Operating Characteristic curve (AUROC)
    \item Maximum F1-Score
    \item Average Precision (AP)
\end{itemize}

These standard metrics allow us to assess the model's performance across all possible decision thresholds, providing a comprehensive evaluation of anomaly detection, especially in highly imbalanced datasets where fixed thresholds may not generalize well.


\subsection{Anomaly Detection Results}\label{sec: anomaly_results}

Our evaluation focuses on detecting anomalies at both the staypoint and agent levels across the NUMOSIM-LA, Urban Anomalies-Berlin, and Urban Anomalies-Atlanta datasets. For agent-level anomaly detection, anomaly scores are computed by taking the maximum score among all associated staypoints for each agent. This approach effectively highlights the most anomalous behavior within each agent’s trajectory, enabling more accurate identification of significant deviations.

\begin{figure*}[htbp]
    \centering
    \begin{subfigure}[b]{0.32\textwidth}
        \centering
        \includegraphics[width=0.9\textwidth]{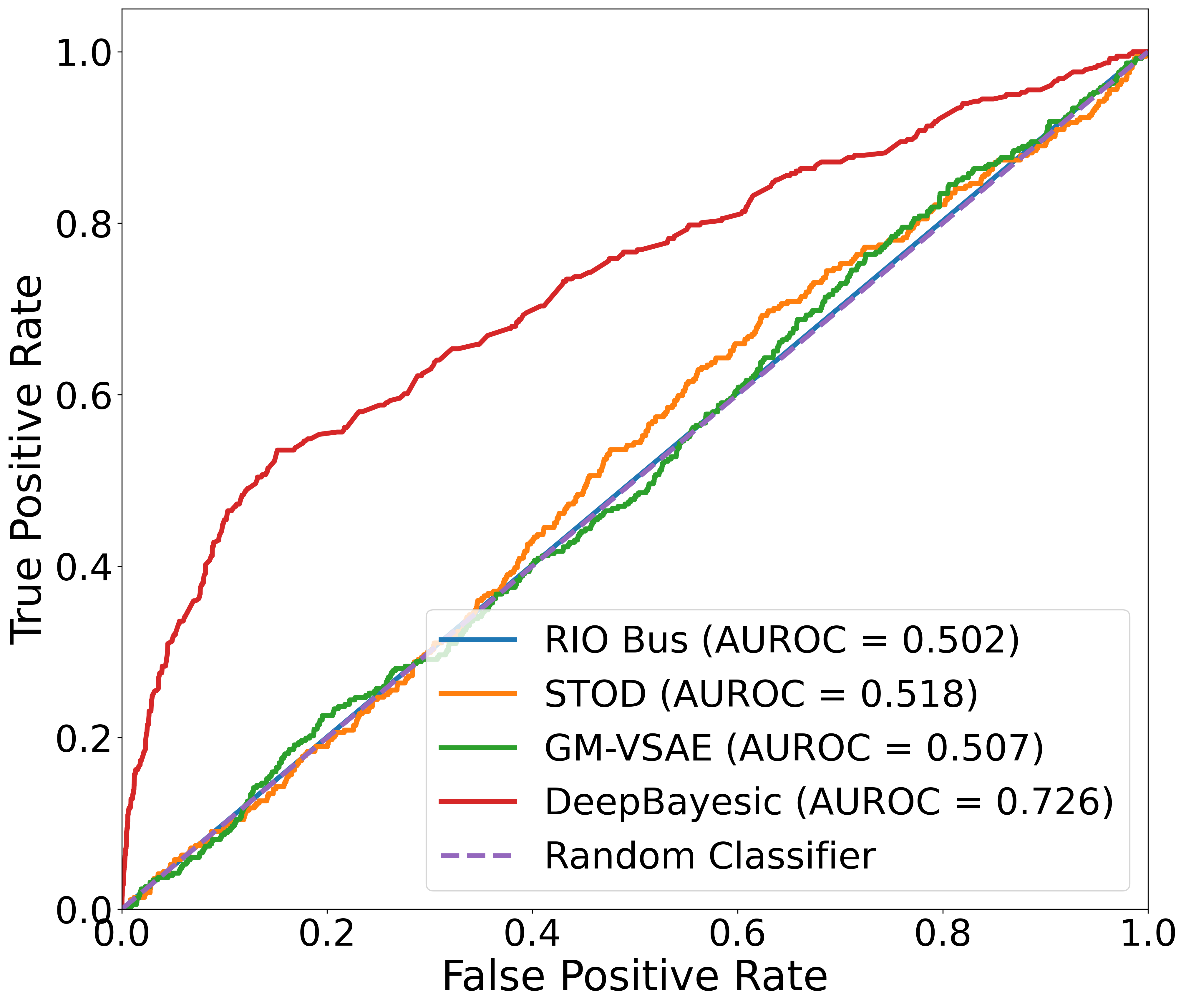}
        \caption{NUMOSIM-LA (Agent)}
        \label{fig:numosim_la_agent}
    \end{subfigure}
    \hfill
    \begin{subfigure}[b]{0.32\textwidth}
        \centering
        \includegraphics[width=0.9\textwidth]{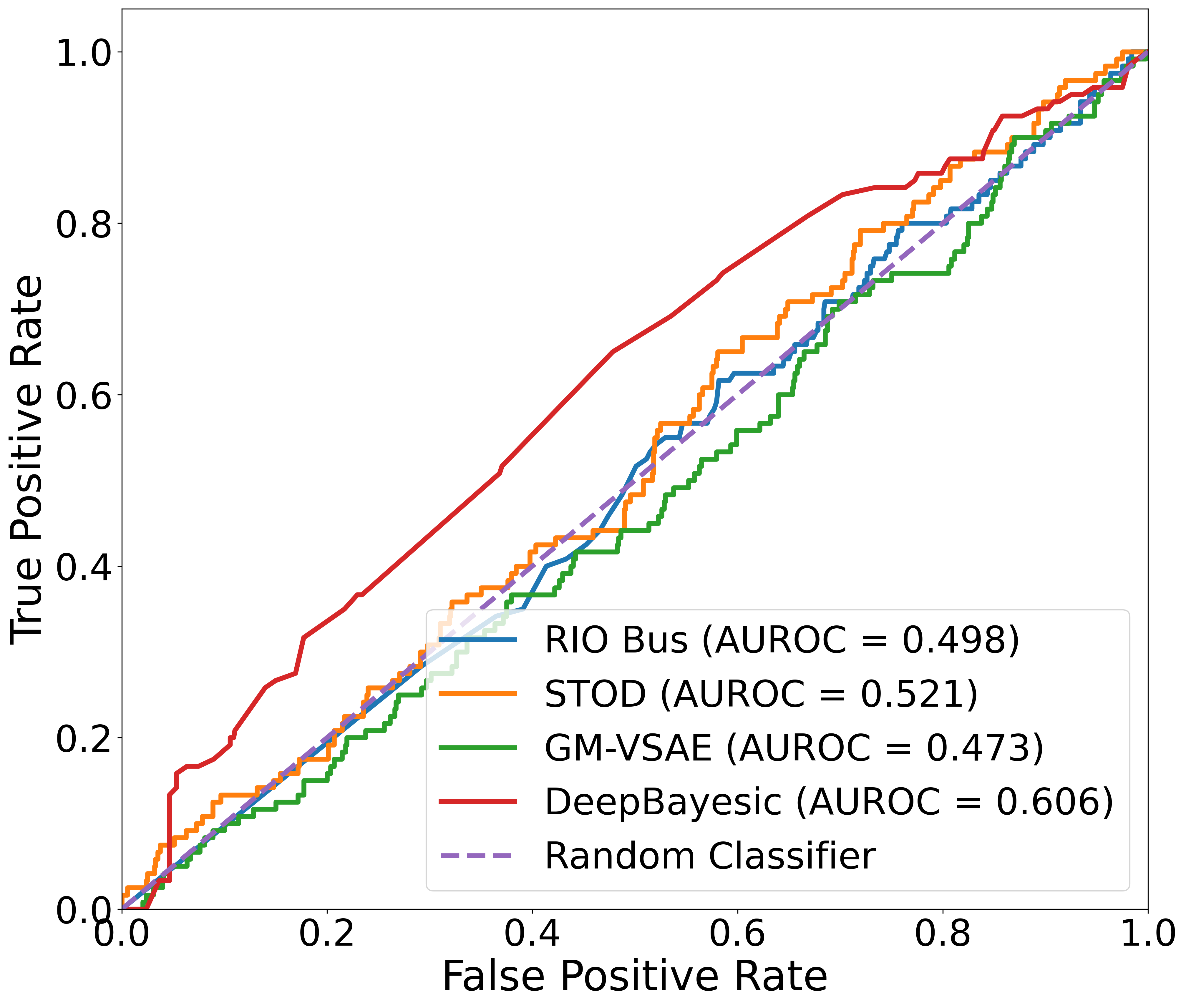}
        \caption{Urban Anomalies-Berlin (Agent)}
        \label{fig:berlin_agent}
    \end{subfigure}
    \hfill
    \begin{subfigure}[b]{0.32\textwidth}
        \centering
        \includegraphics[width=0.9\textwidth]{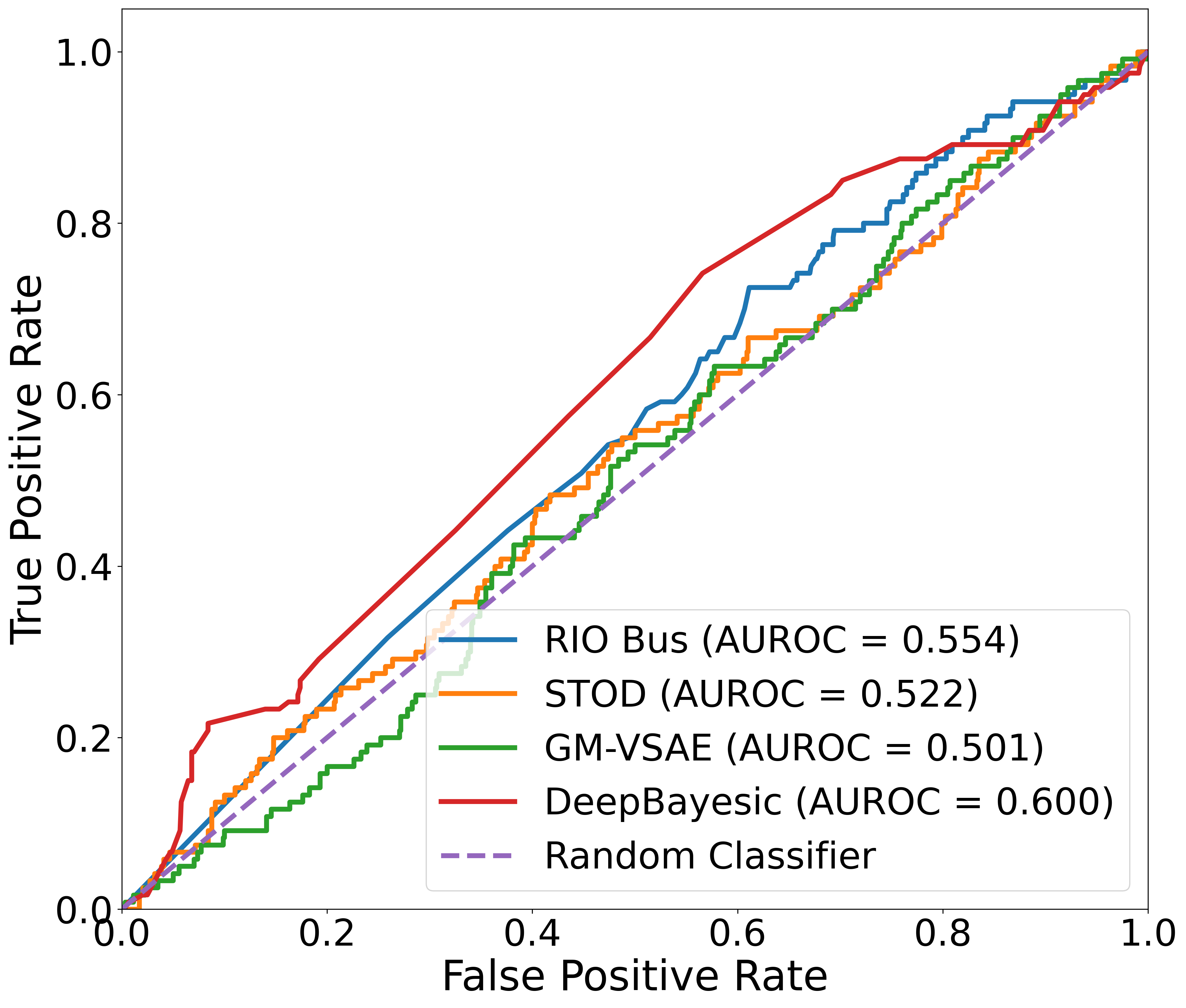}
        \caption{Urban Anomalies-Atlanta (Agent)}
        \label{fig:atlanta_agent}
    \end{subfigure}
    
    \begin{subfigure}[b]{0.32\textwidth}
        \centering
        \includegraphics[width=0.9\textwidth]{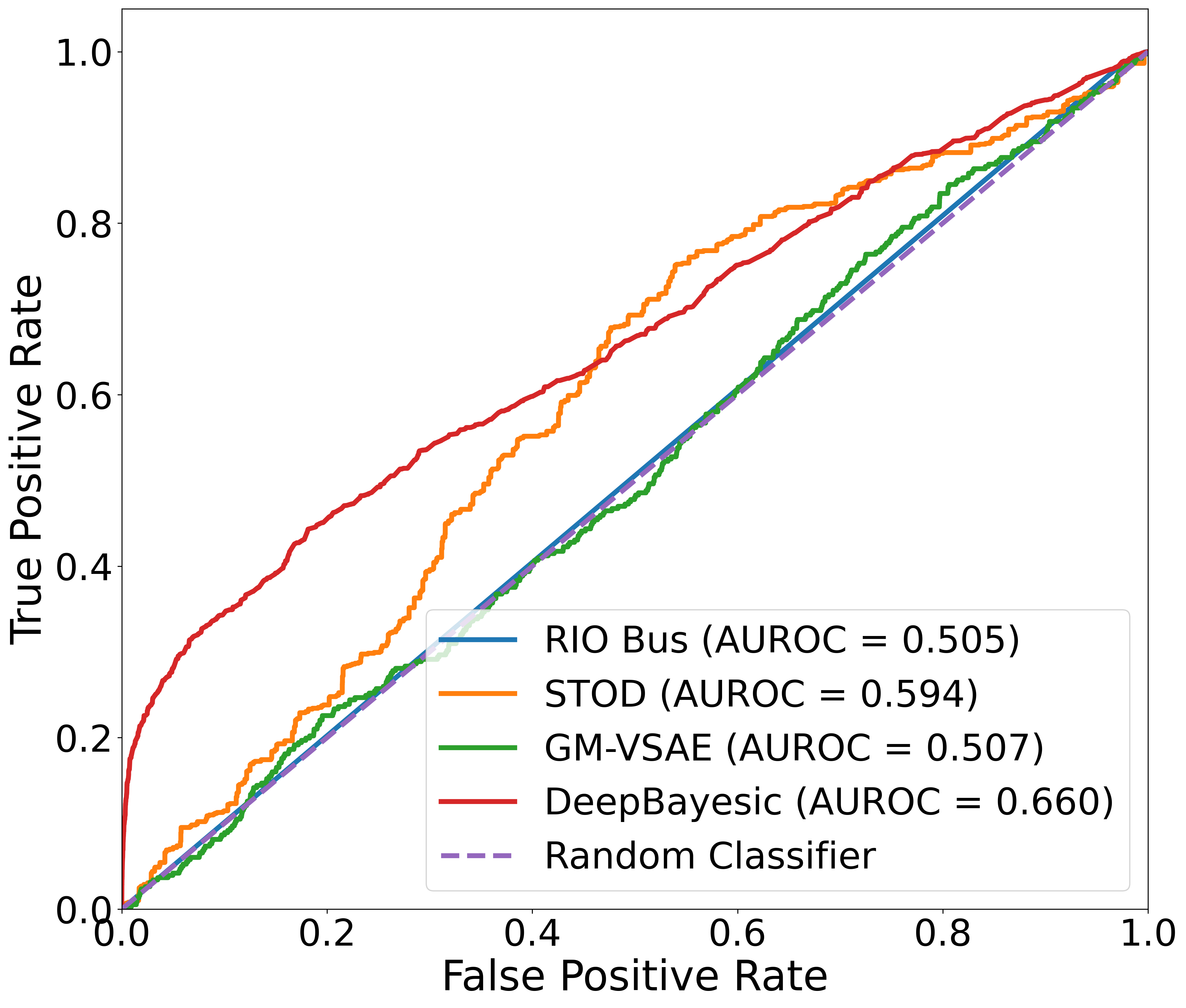}
        \caption{NUMOSIM-LA (Staypoint)}
        \label{fig:numosim_la_staypoint}
    \end{subfigure}
    \hfill
    \begin{subfigure}[b]{0.32\textwidth}
        \centering
        \includegraphics[width=0.9\textwidth]{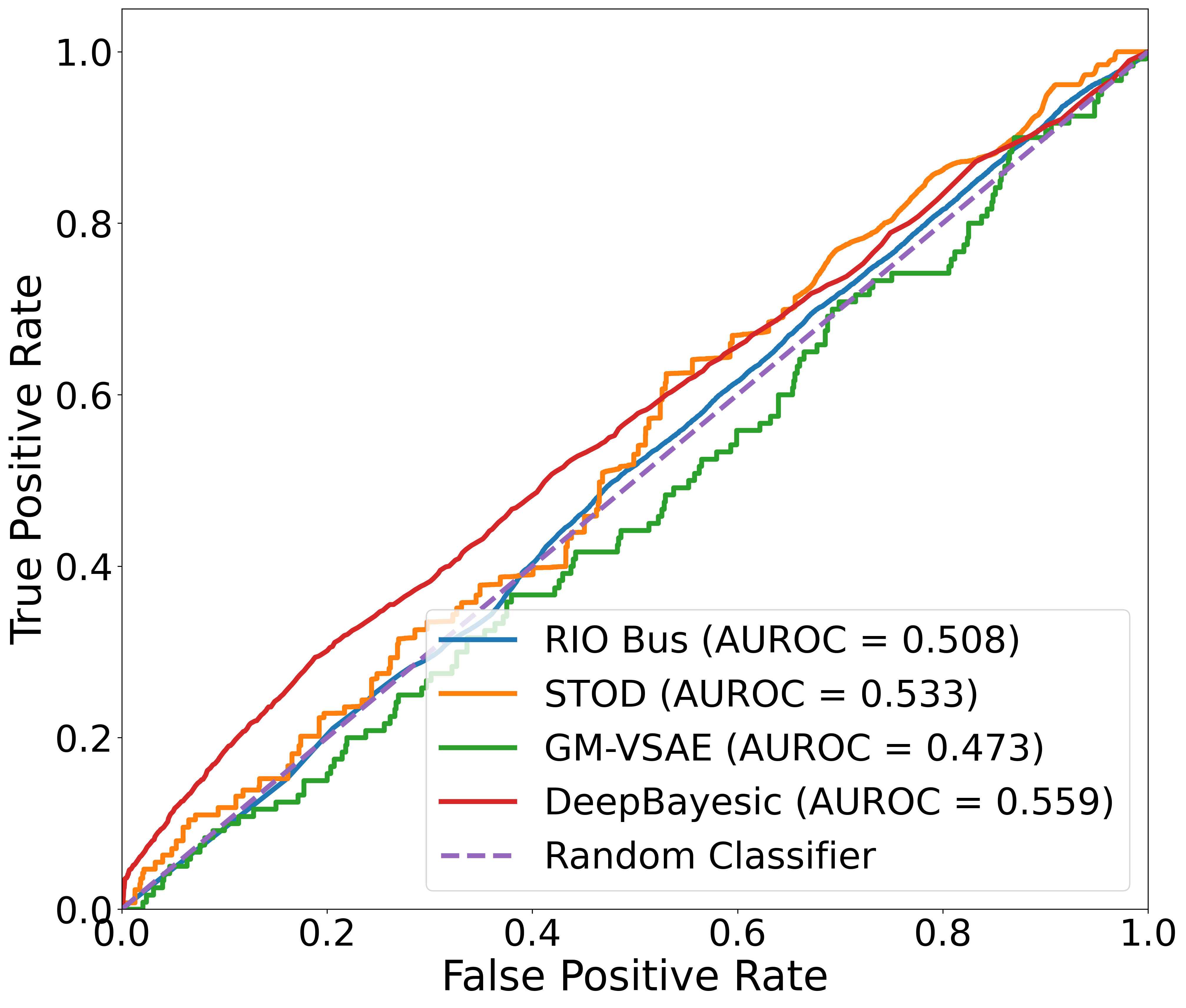}
        \caption{Urban Anomalies-Berlin (Staypoint)}
        \label{fig:berlin_staypoint}
    \end{subfigure}
    \hfill
    \begin{subfigure}[b]{0.32\textwidth}
        \centering
        \includegraphics[width=0.9\textwidth]{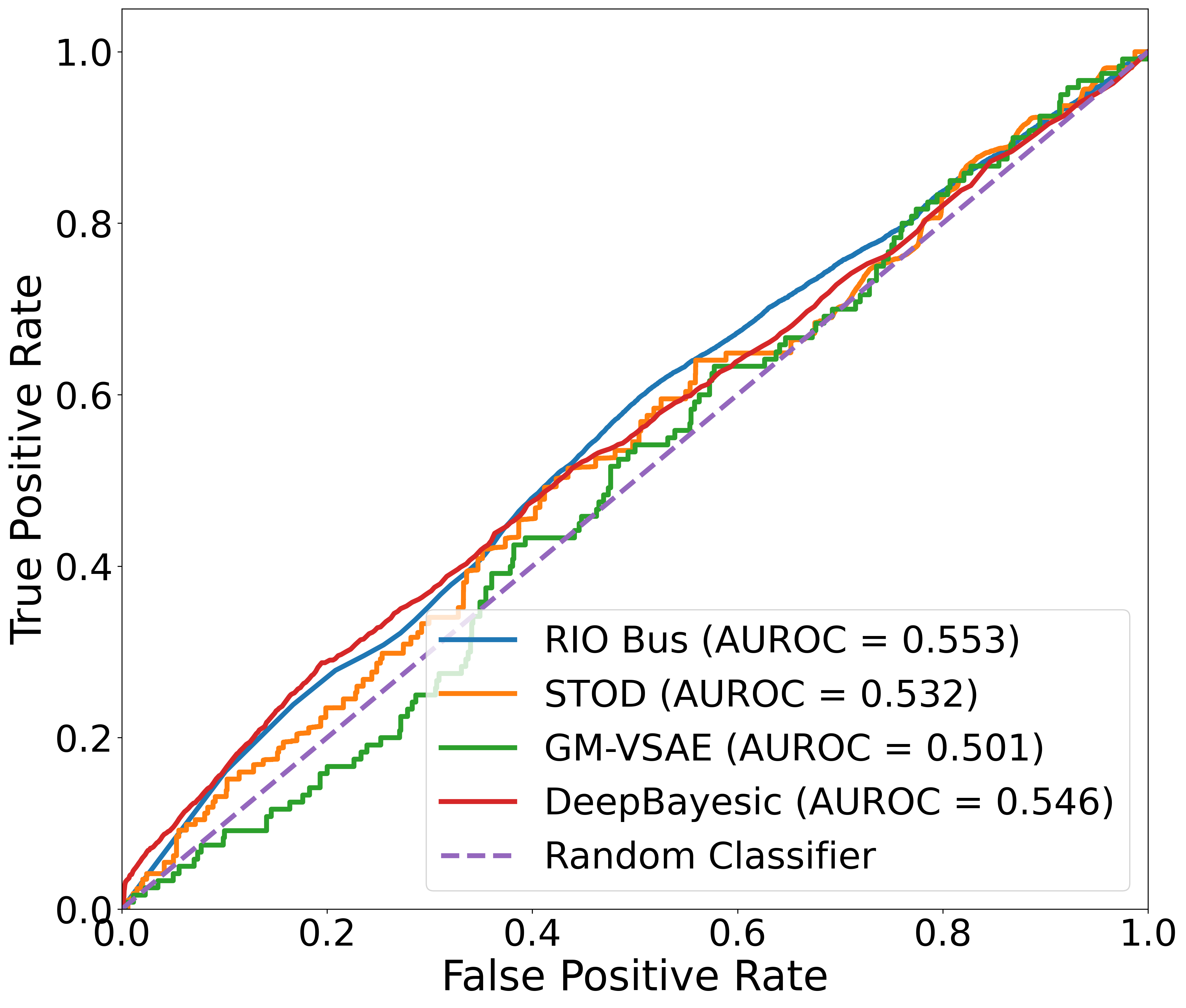}
        \caption{Urban Anomalies-Atlanta (Staypoint)}
        \label{fig:atlanta_staypoint}
    \end{subfigure}
    
    \caption{ROC curves for anomaly detection performance across different datasets and levels of granularity. The top row (a-c) shows the ROC curves for agent-level anomaly detection, while the bottom row (d-f) shows the ROC curves for staypoint-level anomaly detection. Our method, DeepBayesic, is represented in red.}
    \label{fig:roc}
\end{figure*}

We compared the performance of our anomaly detection pipeline against all baseline methods. As shown in Table \ref{tab:results} and the ROC curves in Figure \ref{fig:roc}, our method, DeepBayesic, consistently outperforms the baselines across multiple metrics.

\begin{table*}[htbp]
\centering
\caption{Performance of Our Method and Baseline Methods on NUMOSIM-LA, Urban Anomalies-Berlin, and Urban Anomalies-Atlanta Datasets. The performance metrics (AUPR, AUROC, Precision, and Max F1-Score) are reported for both agent-level and staypoint-level anomaly detection tasks, with the agent-level metrics listed first, followed by the staypoint-level metrics, separated by a `/' within each cell. For GM-VSAE, only agent-level metrics are available.}
\label{tab:results}
\begin{tabular}{|l|l|c|c|c|c|}
\hline
\textbf{Method}                & \textbf{Dataset}        & \textbf{AUPR}  & \textbf{AUROC} & \textbf{Precision} & \textbf{Max F1-Score} \\ \hline
\textbf{DeepBayesic}           & \cellcolor{cyan!15} NUMOSIM-LA              & \cellcolor{cyan!15} \textbf{1.21\%} / \textbf{0.42\%} & \cellcolor{cyan!15} \textbf{72.58\%} / \textbf{65.98\%} & \cellcolor{cyan!15} \textbf{1.91e-03} / 2.11e-04 & \cellcolor{cyan!15} \textbf{5.23e-02} / \textbf{3.74e-02} \\ \cline{2-6} 
                                & \cellcolor{yellow!15} Urban Anomalies-Berlin         & \cellcolor{yellow!15} \textbf{16.54\%} / \textbf{17.71\%} & \cellcolor{yellow!15} \textbf{60.58\%} / \textbf{55.93\%} & \cellcolor{yellow!15} \textbf{1.20e-01} / \textbf{1.28e-01} & \cellcolor{yellow!15} \textbf{2.52e-01} / 2.25e-01 \\ \cline{2-6} 
                                & \cellcolor{green!15} Urban Anomalies-Atlanta        & \cellcolor{green!15} \textbf{16.12\%} / \textbf{16.70\%} & \cellcolor{green!15} \textbf{60.04\%} / 54.64\% & \cellcolor{green!15} 1.20e-01 / 1.26e-01  & \cellcolor{green!15} \textbf{2.52e-01} / 2.23e-01 \\ \hline

\textbf{RIO Bus}               & \cellcolor{cyan!15} NUMOSIM-LA              & \cellcolor{cyan!15} 0.16\% / 0.02\% & \cellcolor{cyan!15} 50.17\% / 50.55\% & \cellcolor{cyan!15} 1.64e-03 / 1.89e-04 & \cellcolor{cyan!15} 3.29e-03 / 3.82e-04 \\ \cline{2-6}
                                & \cellcolor{yellow!15} Urban Anomalies-Berlin         & \cellcolor{yellow!15} 11.91\% / 12.36\% & \cellcolor{yellow!15} 49.81\% / 50.75\% & \cellcolor{yellow!15} 1.20e-01 / 1.23e-01 & \cellcolor{yellow!15} 2.17e-01 / 2.21e-01 \\ \cline{2-6}
                                & \cellcolor{green!15} Urban Anomalies-Atlanta        & \cellcolor{green!15} 13.61\% / 14.87\% & \cellcolor{green!15} 55.36\% / \textbf{55.31\%} & \cellcolor{green!15} 1.20e-01 / 1.25e-01 & \cellcolor{green!15} 2.34e-01 / 2.22e-01\\ \hline

\textbf{STOD}                  & \cellcolor{cyan!15} NUMOSIM-LA              & \cellcolor{cyan!15} 0.19\% / 0.03\% & \cellcolor{cyan!15} 51.81\% / 59.44\% & \cellcolor{cyan!15} 1.82e-03 / \textbf{2.41e-04} & \cellcolor{cyan!15} 4.08e-03 / 3.83e-03 \\ \cline{2-6}
                                & \cellcolor{yellow!15} Urban Anomalies-Berlin         & \cellcolor{yellow!15} 14.64\% / 13.98\% & \cellcolor{yellow!15} 52.11\% / 53.29\% & \cellcolor{yellow!15} 1.16e-01 / 1.22e-01 & \cellcolor{yellow!15} 2.24e-01 / \textbf{2.32e-01} \\ \cline{2-6}
                                & \cellcolor{green!15} Urban Anomalies-Atlanta        & \cellcolor{green!15} 13.02\% / 13.98\% & \cellcolor{green!15} 52.25\% / 53.18\% & \cellcolor{green!15} 1.28e-01 / \textbf{1.37e-01} & \cellcolor{green!15} 2.19e-01 / \textbf{2.29e-01} \\ \hline

\textbf{GM-VSAE}               & \cellcolor{cyan!15} NUMOSIM-LA              & \cellcolor{cyan!15} 0.19\%  & \cellcolor{cyan!15} 50.66\% & \cellcolor{cyan!15} 1.62e-03  & \cellcolor{cyan!15} 4.36e-03  \\ \cline{2-6}
                                & \cellcolor{yellow!15} Urban Anomalies-Berlin         & \cellcolor{yellow!15} 11.28\%  & \cellcolor{yellow!15} 47.29\% & \cellcolor{yellow!15} 9.68e-02 & \cellcolor{yellow!15} 2.18e-01 \\ \cline{2-6}
                                & \cellcolor{green!15} Urban Anomalies-Atlanta        & \cellcolor{green!15} 11.91\% & \cellcolor{green!15} 50.14\% & \cellcolor{green!15} \textbf{1.33e-01}  & \cellcolor{green!15} 2.20e-01 \\ \hline

\end{tabular}
\end{table*}

\textbf{Agent-Level Anomaly Detection:} The ROC curves for agent-level detection, shown in Figure \ref{fig:roc} (a-c), reveal a clear improvement of our method over the baselines, especially on the NUMOSIM-LA dataset. The curves 
indicate that our approach effectively identifies a higher number of true positives while maintaining a low rate of false positives. This superior performance can be attributed, in part, to the integration of personalized agent embeddings, which enable our model to more accurately capture individual behavioral patterns, even with sparse data points. We further explore this hypothesis in an ablation study presented in Section (\ref{sec:ablation}).

\textbf{Staypoint Level Anomaly Detection:} The ROC curves for staypoint-level anomaly detection, depicted in Figure \ref{fig:roc} (d-f), demonstrate that while our method still outperforms the baselines, the overall scores are smaller compared to agent-level detection. This is due to the inherent challenges in detecting anomalies at a finer granularity, where the model must identify deviations in behavior at the level of individual staypoints. Additionally, since agent-level anomaly detection aggregates results from staypoint-level detections, it can mitigate the impact of false positives at the staypoint level.  


\textbf{Max F1-Score and Precision:} As shown in Table \ref{tab:results} 
DeepBayesic consistently achieves the highest AUPR and AUROC scores across the datasets. Notably, on the NUMOSIM-LA dataset, the AUPR scores for the RIOBus and STOD baselines near zero, and their AUROC scores hover around 50\%, suggesting performance close to that of a random classifier. In contrast, DeepBayesic achieves significantly better results.  

Furthermore, DeepBayesic also obtains the highest Max F1-scores at both the agent and staypoint level on the NUMOSIM-LA dataset, while RIOBus and STOD record much lower Max-F1 scores. Although F1-Score and precision are less informative in highly imbalanced datasets such as NUMOSIM-LA and Outliers-Berlin, our model’s performance in these metrics indicates that it outperforms the baseline models.

Overall, our model demonstrates strong performance, especially at the agent level, where the use of personalized embeddings and the DeepBayesic framework enables it to detect subtle anomalies effectively. Our method also remains competitive to the baseline at the staypoint level. 

To better understand the reasons behind DeepBayesic's performance, we further analyzed the prediction outcomes across different agents in Section \ref{sec:personalization} and conducted a comprehensive ablation study in Section \ref{sec:ablation} to examine the contributions of each module to the overall performance .

\subsection{Impact of Personalization on Prediction}\label{sec:personalization}

Personalization plays a central role in enhancing the prediction accuracy of our model. By incorporating personalized agent embeddings, the model captures the unique behavioral patterns of individual agents and tailors its predictions accordingly, thereby improving the detection of subtle anomalies that may be overlooked by a generalized model.

\begin{figure}[htbp]
\centering
\begin{subfigure}{0.45\textwidth}
\centering
\includegraphics[width=\textwidth]{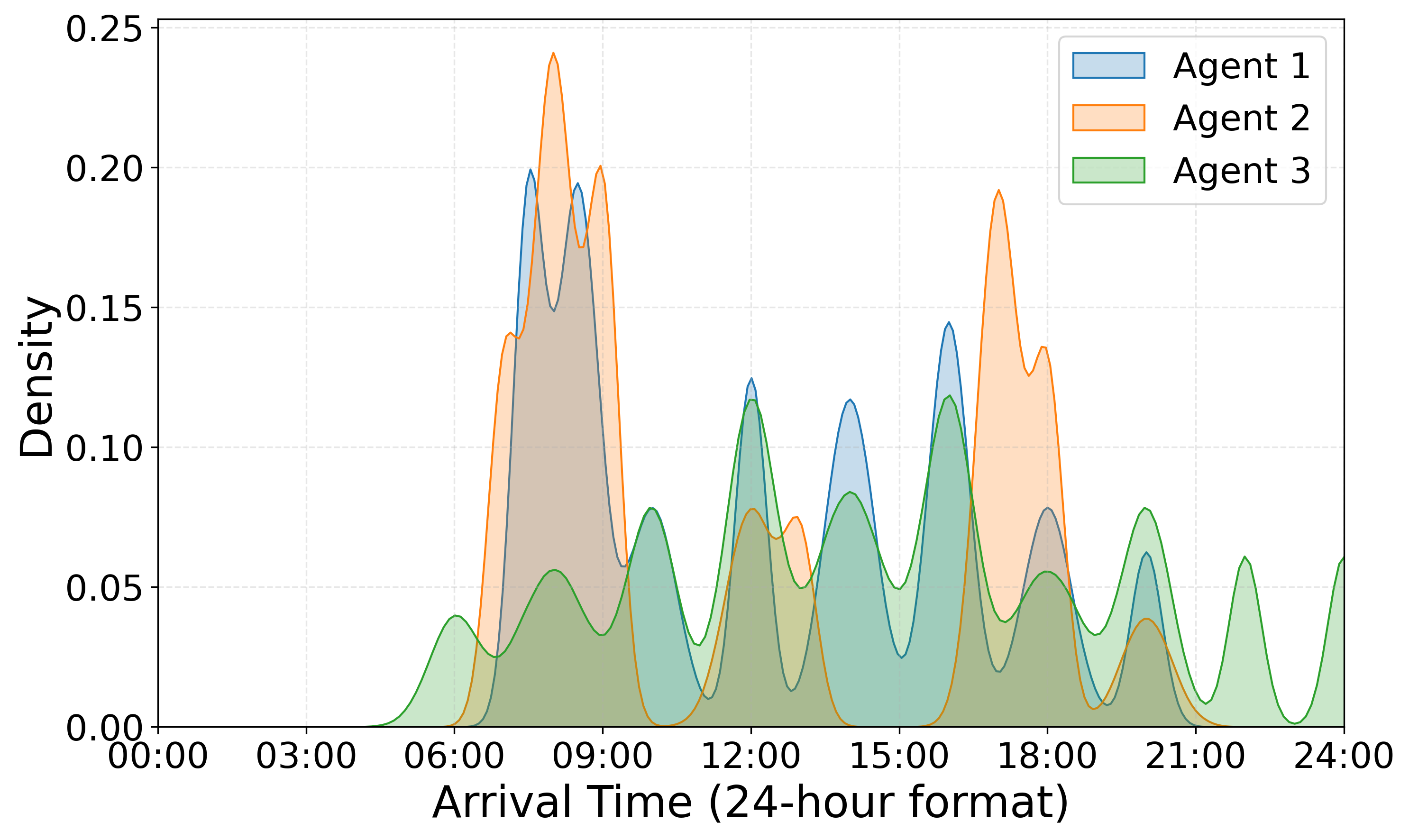}
\caption{Predicted Arrival Time Distribution}
\label{fig:arrival}
\end{subfigure}
\hfill
\begin{subfigure}{0.45\textwidth}
\centering
\includegraphics[width=\textwidth]{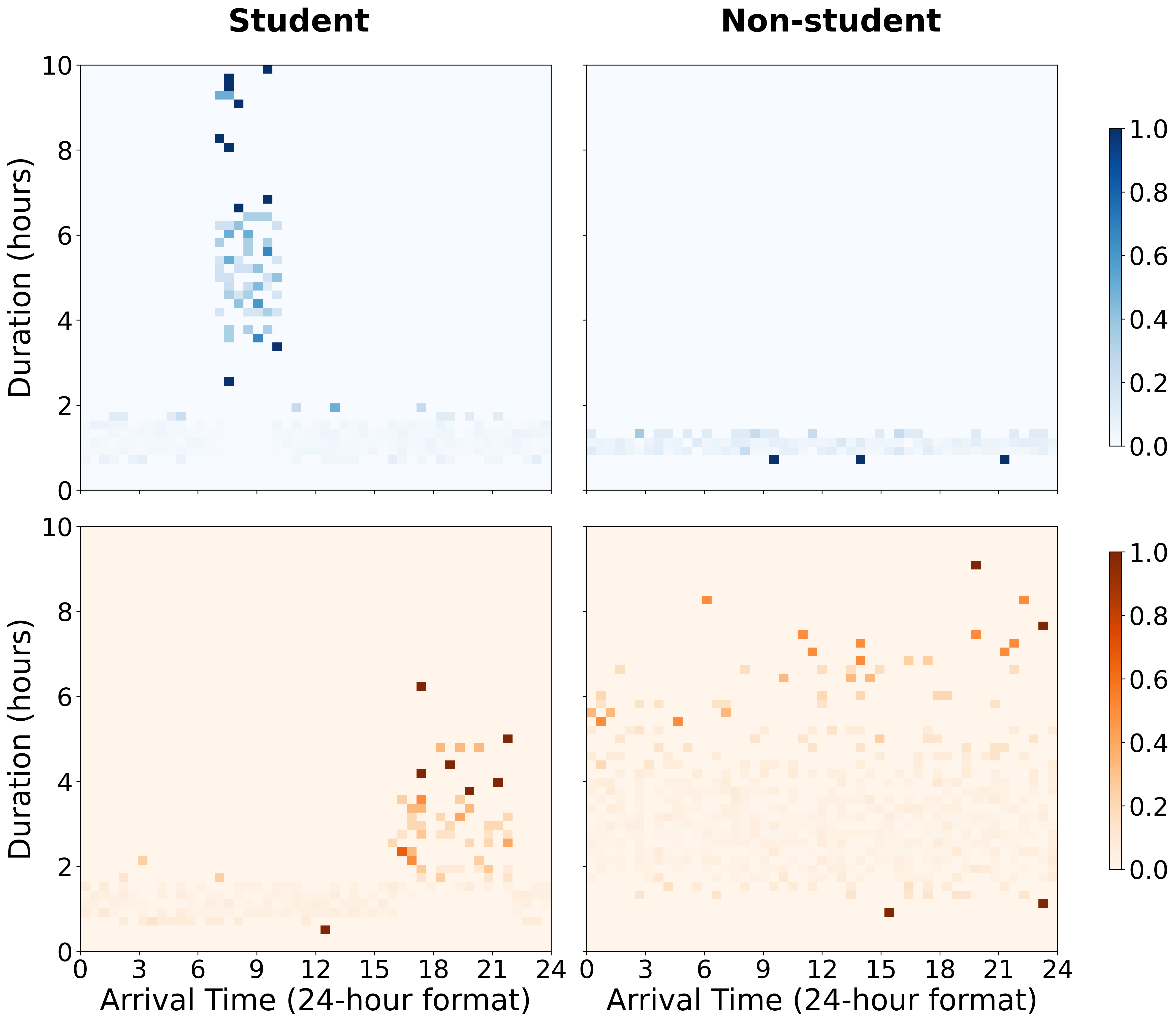}
\caption{Predicted Duration Distribution (Student vs Non-student)}
\label{fig:duration}
\end{subfigure}
\caption{Visualization of (a) the predicted arrival time distribution across multiple agents and (b) the predicted duration distribution conditioned on agent embedding, arrival time, and POI types (blue for school, orange for recreation) for a student agent and a non-student agent.} 
\label{fig:personalization_visualization}
\end{figure}

To assess the effectiveness of personalization, we visualize the predicted conditional distributions of both arrival times and durations for multiple agents. These visualizations, presented in Figures \ref{fig:arrival} and \ref{fig:duration}, highlight the differences in density patterns among agents. 

Figure \ref{fig:arrival} shows the predicted arrival time distributions for three distinct agents. The personalized embeddings capture unique patterns for each agent: Agent 1 (blue) exhibits a bimodal distribution with peaks around 08:00 and 14:00, suggesting a student with morning and afternoon classes; Agent 2 (orange) displays a strong peak at 09:00 and a smaller peak at 18:00, indicative of a typical work schedule; Agent 3 (yellow) shows a more uniform distribution throughout the day, representing an agent with less structured routine.

Figure \ref{fig:duration} provides deeper insight into the model's personalization capabilities by focusing on the duration predictions for two example agents: a student and a non-student. The heatmaps reveal distinct patterns for school and recreation POI types: The student agent typically attends school in the morning, spending 6-7 hours on campus, while the non-student agent rarely visits school and spends significantly less time there. For recreation activities, the student agent usually spends shorter periods in the afternoon and evening, whereas the non-student agent's recreation time is more evenly distributed, indicating a more flexible schedule.

These visualizations confirm that the model effectively differentiates between agents based on their unique patterns, leading to more accurate anomaly detection. For instance, agents who typically follow regular routes and schedules are easily distinguished from those with more erratic behaviors, allowing the model to detect deviations from expected behavior more precisely in both cases.

\subsection{Ablation Study: Impact of Model Components}\label{sec:ablation}

To assess the contribution of each component in our pipeline, we conducted an ablation study by systematically removing one component at a time and evaluating the pipeline's performance using the  metrics introduced in Section \ref{sec:metrics}.

\begin{table*}[htbp]
\centering
\caption{Ablation Study Results for Agent-Level and Staypoint-Level Detection on NUMOSIM-LA and Urban AnomaliesS-BERLIN Datasets. The results for agent-level and staypoint-level are reported together, separated by a `/' within each cell.}
\label{tab:ablation}
\begin{tabular}{|l|l|c|c|}
\hline
\textbf{Configuration} & \textbf{Dataset} & \textbf{AUPR} & \textbf{AUROC} \\ \hline
\textbf{Full Pipeline}          & \cellcolor{cyan!15} NUMOSIM-LA       & \cellcolor{cyan!15} 1.21\% / 0.42\%   & \cellcolor{cyan!15} 72.58\% / 65.98\%  \\ \cline{2-4} 
                       & \cellcolor{yellow!15} Urban Anomalies-BERLIN  & \cellcolor{yellow!15} 16.54\% / 17.71\% & \cellcolor{yellow!15} 60.58\% / 55.93\%  \\ \cline{2-4}
                       & \cellcolor{green!15} Urban Anomalies-ATLANTA & \cellcolor{green!15} 16.12\% / 16.70\% & \cellcolor{green!15} 60.04\% / 54.64\% \\ \hline 
\textbf{Without Arrival Time Prediction} & \cellcolor{cyan!15} NUMOSIM-LA  & \cellcolor{cyan!15} 0.94\% / 0.29\%   & \cellcolor{cyan!15} 70.17\% / 60.87\% \\ \cline{2-4} 
                                & \cellcolor{yellow!15} Urban Anomalies-BERLIN  & \cellcolor{yellow!15} 14.52\% / 12.47\% & \cellcolor{yellow!15} 58.16\% / 54.13\%  \\ \cline{2-4}
                                & \cellcolor{green!15} Urban Anomalies-ATLANTA & \cellcolor{green!15} 10.89\% / 14.25\% & \cellcolor{green!15} 58.12\% / 52.91\% \\ \hline 
\textbf{Without POI Type Prediction}     & \cellcolor{cyan!15} NUMOSIM-LA  & \cellcolor{cyan!15} 0.85\% / 0.31\%   & \cellcolor{cyan!15} 66.82\% / 55.41\%  \\ \cline{2-4} 
                                & \cellcolor{yellow!15} Urban Anomalies-BERLIN  & \cellcolor{yellow!15} 13.37\% / 11.59\% & \cellcolor{yellow!15} 55.62\% / 52.79\%  \\ \cline{2-4} 
                                & \cellcolor{green!15} Urban Anomalies-ATLANTA & \cellcolor{green!15} 0.76\% / 13.02\% & \cellcolor{green!15} 54.74\% / 58.62\% \\ \hline
\textbf{Without Duration Prediction}     & \cellcolor{cyan!15} NUMOSIM-LA  & \cellcolor{cyan!15} 0.76\% / 0.28\%   & \cellcolor{cyan!15} 65.09\% / 51.88\%  \\ \cline{2-4} 
                                & \cellcolor{yellow!15} Urban Anomalies-BERLIN  & \cellcolor{yellow!15} 12.08\% / 10.42\% & \cellcolor{yellow!15} 52.94\% / 51.03\%  \\ \cline{2-4} 
                                & \cellcolor{green!15} Urban Anomalies-ATLANTA & \cellcolor{green!15} 8.66\% / 12.50\% & \cellcolor{green!15} 53.42\% / 52.15\% \\ \hline
\textbf{Without Agent Embedding}         & \cellcolor{cyan!15} NUMOSIM-LA  & \cellcolor{cyan!15} 0.01\% / 0.01\%   & \cellcolor{cyan!15} 50.88\% / 52.32\%  \\ \cline{2-4} 
                                & \cellcolor{yellow!15} Urban Anomalies-BERLIN  & \cellcolor{yellow!15} 10.47\% / 8.19\%  & \cellcolor{yellow!15} 49.83\% / 52.69\%  \\ \cline{2-4} 
                                & \cellcolor{green!15} Urban Anomalies-ATLANTA & \cellcolor{green!15} 6.43\% / 10.11\% & \cellcolor{green!15} 51.97\% / 50.83\% \\ \hline
\end{tabular}
\end{table*}

\begin{table*}[h]
\centering
\caption{Results of DeepBayesic + Visit Rate Incorporation on Agent-Level Detection across NUMOSIM-LA, Urban Anomalies-Berlin, and Urban Anomalies-Atlanta Datasets. The performance metrics (AUPR, AUROC, Precision, and Max F1 Score) are reported for agent-level anomaly detection tasks.}
\label{tab:visit_rate_results}
\begin{tabular}{|l|l|c|c|c|c|}
\hline
\textbf{Dataset} & \textbf{Method} & \textbf{AUPR}  & \textbf{AUROC} & \textbf{Precision} & \textbf{Max F1 Score} \\ \hline

\multirow{2}{*}{\textbf{NUMOSIM-LA}} 
& \textit{Visit Rate Baseline \cite{stanford2024numosimsyntheticmobilitydataset}} & 1.64\% & 64.6\% & - & - \\ \cline{2-6}
& \textbf{DeepBayesic + Visit Rate} & \cellcolor{cyan!15} 5.81\% & \cellcolor{cyan!15} 66.17\% & \cellcolor{cyan!15} 4.34e-01 & \cellcolor{cyan!15} 1.63e-01 \\ \hline

\textbf{Urban Anomalies-Berlin} 
& \textbf{DeepBayesic + Visit Rate} & \cellcolor{yellow!15} 36.37\% & \cellcolor{yellow!15} 71.32\% & \cellcolor{yellow!15} 1.00e+00 & \cellcolor{yellow!15} 3.83e-01 \\ \hline

\textbf{Urban Anomalies-Atlanta} 
& \textbf{DeepBayesic + Visit Rate} & \cellcolor{green!15} 41.38\% & \cellcolor{green!15} 77.38\% & \cellcolor{green!15} 9.17e-01 & \cellcolor{green!15} 4.53e-01 \\ \hline

\end{tabular}
\end{table*}

Table \ref{tab:ablation} summarizes the results of this ablation study for both NUMOSIM-LA and Urban Anomalies datasets, highlighting the performance degradation observed when each component is excluded from the pipeline. The results show that the most significant drop in performance occurs when the agent embedding is removed: the agent-level AUPR decreases from 1.21\% to 0.01\% on the NUMOSIM-LA dataset, from 16.54\% to 10.47\% on the Urban Anomalies-Berlin dataset, and from 16.12\% to 6.43\% on the Urban Anomalies-Atlanta dataset. Similar declines are observed across other metrics, underscoring the critical role of agent embedding in model performance. 

Additionally, removing the arrival time estimation model, the POI type estimation model, or the duration estimation model also leads to a decrease in performance, indicating that each of these components contributes significantly to the overall effectiveness of the model.


\subsection{Visit Rate Analysis}\label{sec:visit_rate}
The NUMOSIM paper \cite{stanford2024numosimsyntheticmobilitydataset} demonstrated that a simple visit rate model --- tracking the frequency of visits to various points of interest (POIs) --- can outperform many existing baselines in detecting anomalies. We chose not to incorporate the visit rate directly into our main model to ensure a fair comparison with other baseline methods that do not utilize this additional knowledge. This approach allows the evaluation and comparisons in Section \ref{sec: anomaly_results} to focus solely on the core concepts presented in the paper, rather than on specific attributes. 

However, for completeness, we integrated the visit rate attribute into our model and compared its performance against the baseline provided in the original NUMOSIM paper. The visit rate model computes anomaly scores by comparing the observed visit rate between the training and testing periods, normalized by the standard deviation observed during training. To integrate this into our model, we normalize the computed anomaly scores between 0 and 1 and multiply them with the final anomaly scores produced by our framework, ensuring that the visit rate contributes proportionally to the overall anomaly score. 

The impact of incorporating the visit rate into our framework is summarized in Table \ref{tab:visit_rate_results}, where we report the performance metrics for the NUMOSIM-LA, Urban Anomalies-Berlin, and Urban Anomalies-Atlanta datasets. The table includes only AUPR and AUROC for the baseline visit rate model as these are the only metrics provided in the original paper. Notably, our approach significantly surpass the Visit Rate Baseline on agent level on NUMOSIM-LA dataset. It also surpasses DeepBayesic (without incorporating visit rate) on Urban Anomalies-Berlin and Urban Anomalies-Atlanta datasets.  These results highlight its effectiveness in detecting subtle anomalies.


%% file: conclusion.tex
\section{Conclusion}\label{sec:conclusion}

In conclusion, DeepBayesic represents a return to foundational principles, demonstrating that by combining the strengths of Bayesian theory with advanced neural density estimation techniques, we can develop powerful, interpretable, and effective solutions for spatiotemporal anomaly detection. Our approach also highlights the importance of personalized modeling in capturing the unique behavioral patterns of individuals in mobility data. By incorporating personalization through learned agent embeddings, the model is able to detect subtle and context-specific anomalies, even in sparse datasets. This integrated approach ensures both robustness and accuracy while providing a solid foundation for future enhancements. 